\documentclass[11pt]{article}
\usepackage{acl2012}
\usepackage{setspace}
\usepackage{times}
\usepackage{latexsym}
\usepackage{amsmath}
\usepackage{multirow}
\usepackage{url}
\usepackage{graphicx}
\usepackage{epstopdf}
\usepackage{enumitem}
\usepackage{algorithmic}
\usepackage[lined,boxed,commentsnumbered]{algorithm2e}
\usepackage[small]{caption}

\DeclareMathOperator*{\argmin}{arg\,min}
\usepackage[small,compact]{titlesec}
\newenvironment{enum}{
\begin{enumerate}
  \scriptsize
    \setlength{\itemsep}{1pt}
    \setlength{\parskip}{0pt}
    \setlength{\parsep}{0pt}
    \setlength{\itemindent}{0cm}

}{\end{enumerate}}

\title{Unsupervised Topic Modeling Approaches to Decision Summarization in Spoken Meetings}

\author{Lu Wang \\
  Department of Computer Science \\
  Cornell University \\
  Ithaca, NY 14853 \\
  {\tt luwang@cs.cornell.edu} \\\And
  Claire Cardie \\
  Department of Computer Science \\
  Cornell University \\
  Ithaca, NY 14853 \\
  {\tt cardie@cs.cornell.edu} \\}
  
\begin{document}

\maketitle

\vspace*{-0.5in}

\begin{abstract}
We present a token-level decision summarization framework that utilizes the latent topic structures of utterances to identify ``summary-worthy" words. Concretely, a series of unsupervised topic models is explored and experimental results show that fine-grained topic models, which discover topics at the utterance-level rather than the document-level, can better identify the gist of the decision-making process. Moreover, our proposed token-level summarization approach, which is able to remove redundancies within utterances, outperforms existing utterance ranking based summarization methods. Finally, context information is also investigated to add additional relevant information to the summary.
%We present a word- and phrase- level summarization framework by using unsupervised topic modeling approaches. A progressing of unsupervised topic model based approaches for decision summarization in spoken meetings is explored and the experimental results show that fine-grained topic models can better identify the gist information in the decision-making process. We also find that token-level summarization can remove more redundancy than utterance-level summarization for meetings. In addition, adding context information to token-level summarization methods performs comparable to supervised learning approaches in terms of ROUGE score.
\end{abstract}

\section{Introduction}
\label{intro}
Meetings are an important way for information sharing and collaboration, where people can discuss problems and make concrete decisions. Not surprisingly, there is an increasing interest in developing methods for extractive summarization for meetings and conversations \cite{Zechner:2002:ASO:638178.638181,Maskey2005,Galley:2006:SCR:1610075.1610126,Lin:2010:RMF:1858681.1858690,Murray:2010:ITA:1857999.1858131}.
\newcite{MurrayEtAl:book:2011} describe the specific need for {\em focused summaries} of meetings, i.e., summaries of a particular aspect of a meeting rather than of the meeting as a whole. For example, the decisions made, the action items that emerged and the problems arised are all important outcomes of meetings. In particular, decision summaries would allow participants to review decisions from previous meetings and understand the related topics quickly, which facilitates preparation for the upcoming meetings.
%NEW CHANGE
%For non-attendants, they can go through the meeting summaries to get up to date on the group discussions.

%Our goal is to provide concise and informative decision summaries of meetings, thus saving the users from browsing through the whole meeting. Mostly, text summarization techniques are applied on speech documents and some special characteristics of speech are taken into account~\cite{Zechner:2002:ASO:638178.638181},~\cite{Buist04automaticsummarization},~\cite{Lin2010}. However, the characteristics of speech, such as ungrammatical structure, disfluencies, high word error rate in the transcripts, makes the task of speech summarization in need of removing more redundancy. 

%We aim to produce topically coherent and non-redundant summaries. Unsupervised topic modeling approaches are desirable for this task due to several reasons. Firstly, with the diversity of meeting content, supervised methods which require human written summaries are not often practical and the framework must be robust and flexible with regard to domain. Another reason is the characteristics of speech, including disfluencies, high word error rates, absence of punctuation, interruptions and hesitations. These factors are problematic for methods replying on manually constructed knowledge resources, as they may not tackle every problem caused by speech. Furthermore, some other challenges are presented in Table 1 for decision summarization which will be investigated in this paper.

\begin{figure}[t]

    %\hspace{-1.5cm}
    {\footnotesize
    \setlength{\baselineskip}{0pt}
  
    \begin{tabular}{|l|}
    \hline
        
    A:We decided our target group is the focus on who can\\ afford it , (1)\\
    %B:what do people think about this kinetic battery idea ? (5)\\

    %A:[vocalsound] I think it's awesome . I think it's really cool . (5)\\
    {\bf B:Uh I'm kinda liking the idea of latex , if if spongy is}\\ {\bf the in thing . (2)}\\
    {\bf B:what I've seen , just not related to this , but of latex}\\ {\bf cases before , is that [vocalsound] there's uh like a hard}\\ {\bf plastic inside , and it's just covered with the latex . (2)}\\
    C:Um [disfmarker] And I think if we wanna keep our costs\\ down , we should just go for pushbuttons , (3)\\
    %B:I think it might be better to put our money into the stuff like the\\ kinetic battery and the cool case (5)\\
    {\bf D:but if it's gonna be in a latex type thing and that's}\\ {\bf gonna look cool , then that's probably gonna have a}\\ {\bf bigger impact than the scroll wheel . (2)}\\
    A:we're gonna go with um type pushbuttons , (3)\\
    A:So we're gonna have like a menu button , (4)\\
    C:uh volume , favourite channels , uh and menu . (4)\\
    A:Pre-set channels (4)\\~\\
    
{\bf Decision Abstracts (Summary)}\\
{\sc Decision 1}: The target group comprises of individuals\\ who can afford the product.\\
%{\sc Decision 2}: The remote will use a kinetic battery.\\
{\sc Decision 2}: The remote will have a latex case.\\
{\sc Decision 3}: The remote will have pushbuttons.\\
{\sc Decision 4}: The remote will have a power button, volume\\ buttons, channel preset buttons, and a menu button.\\
        \hline
    \end{tabular}
    }
    \vspace{-0.3cm}
    \caption{\footnotesize A clip of a meeting from the AMI meeting corpus~\cite{Carletta05theami}. A, B, C and D refer to distinct speakers; the numbers in parentheses indicate the associated meeting decision: {\sc decision 1}, {\sc 2}, {\sc 3} or {\sc 4}. Also shown is the gold-standard (manual) abstract (summary) for each decision.}
    \label{fig:example}
\end{figure}
\vspace{-0.7cm}

% NEW CHANGE
Meeting conversation is intrinsically different from well-written text, as meetings may not be well organized and most utterances have low density of salient content. Therefore, multiple problems need to be addressed for speech summarization. Consider the sample dialogue snippet in Figure~\ref{fig:example} from the AMI meeting corpus \cite{Carletta05theami}.  Only {\it decision-related dialogue acts (DRDAs)} --- utterances  at least one decision made in the meeting\footnote{These DRDAs are annotated in the AMI corpus and usually contain the decision content. They are similar, but not completely equivalent, to the {\it decision dialogue acts (DDAs)} of~\newcite{Bui:2009:EDM:1708376.1708410},~\newcite{Fernandez},~\newcite{Frampton:2009:RDD:1699648.1699659}.} --- are listed and ordered by time. Each DRDA is labeled numerically according to the decision it supports; so the second and third utterances (in {\bf bold}) support {\sc Decision 2}, as do the fifth utterance in the snippet. Manually constructed {\it decision abstracts} for each decision are shown at the bottom of the figure. 

Besides the prevalent dialogue phenomena (such as ``Uh I'm kinda liking" in Figure~\ref{fig:example}), disfluencies and off-topic expressions, we notice that single utterance is usually not informative enough to form a decision. For instance, no single DRDA associated with {\sc Decision 4} corresponds all that well with its decision abstract: ``pushbuttons", ``menu button" and ``Pre-set channels" are mentioned in separate DAs. As a result, extractive summarization methods that select individual utterance to form the summary will perform poorly. 

Furthermore, it is difficult to identify the core topic when multiple topics are discussed in one utterance. For example, all of the bold DRDAs supporting {\sc Decision 2} contain the word ``latex". However, the last DA in bold also mentions ``bigger impact" and ``the scroll wheel'', which are not specifically relevant for {\sc Decision 2}. %though they can be crucial for other decisions. 
Though this problem can be approached by training a classifier to identify the relevant phrases and ignore the irrelevant ones or dialogue phenomena, it needs expensive human annotation and is limited to the specific domain.

%Notice that all of the three DRDAs in bold supporting {\sc Decision 2} (``The remote will have a latex case.") contain the ``latex" concept which increases the confidence of taking ``latex" into summary. However, the last DA in bold discusses not only ``the latex type thing", but also ``bigger impact" and ``the scroll wheel'', which are not very relevant for this decision, though they can be crucial for other decisions in the meeting. So identifying the core topic for each decision is important. Though this problem can be approached by training a classifer to identify the relevant phrases and ignore the irrelevant ones and dialogue phenomena (such as ``Uh I'm kinda liking" in the second DRDA in Figure~\ref{fig:example}), it needs expensive human annotation and is limited to the domain the classifier is trained on.

Note also that for {\sc Decision 4}, the ``power button" is not specified in any of the listed DRDAs supporting it. By looking at the transcript, we find ``power button" mentioned in one of the preceding, but not decision-related DAs. Consequently another challenge would be to add complementary knowledge when the DRDAs cannot provide complete information. 

%In addtion, some of the utterances are ill-structured and incomplete, where the disfluencies or interruptions make the expressions ungrammatical. 
%Therefore, we need a summarization approach that is not only tolerant to speech-related characteristics, but can also determine the key semantic content, capture potential multi-topics for a single decision, leverage available context information and be easily transferable between domains. %Unsupervised methods are preferred for this task as they do not depend on expensive manually annotated domain-specific traning data.
%We believe that summarizing at the granularity level of meetings is not sufficient and can be significantly refined. 
Therefore, we need a summarization approach that is tolerant of dialogue phenomena, can determine the key semantic content and is easily transferable between domains. Recently, topic modeling approaches have been investigated and achieved state-of-the-art results in multi-document summarization~\cite{Haghighi:2009:ECM:1620754.1620807,Celikyilmaz:2010:HHM:1858681.1858765}. Thus, topic models appear to better ref for document similarity w.r.t. semantic concepts than simple literal word matching. However, very little work has investigated its role in spoken document summarization~\cite{Chen:2008:ESD:1332129.1332195,DBLP:conf/interspeech/Hazen11}, and much less conducted comparisons among topic modeling approaches for focused summarization in meetings.%~\cite{MurrayEtAl:book:2011}.

% NEW CHANGE
%In addition, \newcite{murray:generating} show that users much prefer {\em abstractive summaries} over extracts when the text to be summarized is a conversation.
In contrast to previous work, %that extracts a small set of the most informative utterances as the summary, %In this paper, we are interested token-level meeting summarization approaches and investigate unsupervised topic modeling methods to identify summary-worthy words in decision-making process to make up decision summaries. 
we study the unsupervised token-level decision summarization in meetings by identifying a concise set of key words or phrases, which can either be output as a compact summary or be a starting point to generate abstractive summaries. This paper addresses problems mentioned above and make contributions as follows:

\begin{itemize}[leftmargin=0.3cm]
\vspace{-0.2cm}
\item
%%%ctc: uh-oh, i thought the methods above were token-level.
As a step towards creating the abstractive summaries that people prefer when dealing with spoken language~\cite{murray:generating}, we propose a token-level rather than sentence-level framework for identifying components of the summary. %In contrast to previous work either uses whole utterances as summaries, we propose a token-level summarization framework and compare it with the existing sentence ranking algorithms that output entire dialogue acts (DAs) as extractive summaries. 
Experimental results show that, compared to the sentence ranking based summarization algorithms, our token-level summarization framework can better identify the summary-worthy words and remove the redundancies.% caused by the disfluency, ungrammatical structure and errors in the speech as well. 
%Based on the topic models, we exploit numerious summarization metrics for characterizing words' or DAs' candidacy for inclusion in summaries. The experimental results demonstrate that our token-level summarization metrics can produce better summaries and is easier to leverage context information.

\vspace{-0.2cm}
\item
Rather than employing supervised learning methods that rely on costly manual annotation, we explore and evaluate topic modeling approaches of different granularities for the unsupervised decision summarization at both the token-level and dialogue act-level.
%By assuming all decision-related dialogue acts (DRDAs) have been identified, we intend to use topic modeling approaches to discover the topic structure within DAs, catch the crux information for each decision and produce a summary in the form of concise {\it decision abstracts} (see Table 1), one for each decision made. 
%We evaluate the methods (using the AMI corpus~\cite{Carletta05theami}) under two input settings --- in the {\bf True Clusterings} setting, we assume that the DRDAs for each meeting have been perfectly grouped according to the decision(s) each supports; in the {\bf System Clusterings} setting, an automated system performs the DRDA-decision pairing. 
%%%ctc: should this next line be "token-level"?  i thought so, but based on the following paragraph, 
%%%     maybe it's supposed to be DA-level?
We investigate three topic models --- Local LDA (LocalLDA)~\cite{Brody:2010:UAM:1857999.1858121}, Multi-grain LDA (MG-LDA)~\cite{Titov:2008:MOR:1367497.1367513} and Segmented Topic Model (STM)~\cite{Du:2010:STM:1842816.1842821} --- which can utilize the latent topic structure on utterance level instead of document level. Under our proposed token-level summarization framework, three fine-grained models outperform the basic LDA model and two extractive baselines that select the longest and the most representative utterance for each decision, respectively. (ROUGE-SU4 F score of 14.82\% for STM vs. 13.58\% and 13.46\% for the baselines, given the perfect clusterings of DRDAs.)

%(ROUGE-SU4 F score of 14.82\% for STM vs. 13.58\% and 13.46\% for the baselines, given the True Clusterings of DRDAs.)

%(ROUGE-1 F score of 37.32\% for STM vs. 32.61\% and 33.32\% for the baselines; ROUGE-SU4 F score of 14.82\% for STM vs. 13.58\% and 13.46\% for the baselines, given the True Clusterings of DRDAs.) Furthermore, the approaches also perform comparably to two supervised learning methods (ROUGE-1 F scores of 35.53\% and 40.39\%; ROUGE-SU4 F scores of 14.03\% and 16.24\%).
%We compare three fine-grained topic models (Local LDA (LocalLDA)~\cite{Brody:2010:UAM:1857999.1858121}, Multi-grain LDA (MG-LDA)~\cite{Titov:2008:MOR:1367497.1367513} and Segmented Topic Model (STM)~\cite{Du:2010:STM:1842816.1842821}) to basic topci models. 
\vspace{-0.2cm}
\item
In line with prior research that explore the role of context for utterance-based extractive summarization~\cite{DBLP:conf/interspeech/MurrayR07}, we investigate the role of context in our token-level summarization framework. %Finally, we also explore methods incorporating token-level contextual information to improve DA-based summaries. 
%For a given DRDA (denoted as {\it center DA}), two types of context are investigated --- adjacent DAs and the most similar DAs in terms of TF-IDF score. We find that the two types of context have comparable effect on the summary, but selecting words from the dominant topic of the center DA for the cluster have better F scores than selecting words from the dominant topic of each DA itself (For ROUGE-1 f scores: 28.62\% vs. 18.46\%, using adjacent context; 21.55\%  vs. 27.19\%, using
%ctc: it doesn't make sense that the recall can exceed the upperbound recall
%TF-IDF context). Moreover, by leveraging context, the recall exceeds the provided upperbound's recall (ROUGE-1 recall: 48.10\% vs. 45.05\% for upperbound by only using DRDA) although the F score decreases after adding context information. Finally, we show that when the true DRDA clusterings are unknown, adding context can improve both the recall and F score.
For the given clusters of DRDAs, We study two types of context information --- the DAs preceding and succeeding a DRDA and DAs of high TF-IDF similarity with a DRDA. We also investigate two ways to select relevant words from the context DA. Experimental results show that two types of context have comparable effect, but selecting words from the dominant topic of the center DRDA performs better than from the dominant topic of the context DA. Moreover, by leveraging context, the recall exceeds the provided upperbound's recall (ROUGE-1 recall: 48.10\% vs. 45.05\% for upperbound by using DRDA only) although the F scores decrease after adding context information. Finally, we show that when the true DRDA clusterings are not available, adding context can improve both the recall and F score.

\end{itemize}

%This paper is structured as follows. After an overview of related work, we begin with introducing our token-level summarization framework built on topic models, and several existing sentence scoring metrics for utterance-level summarization for comparison. Then we briefly describe three fine-grained topic models. Experimental setup and results are shown and analyzed in Section 5 and 6. %We conclude with a discussion of the effectiveness of our proposed unsupervised summarization framework, and the potential use for abstractive summaries generation.

​

%In this paper, we focus on generic unsupervised meeting summarization methods which can be used as the basis for further summarization tasks. Topic models comparison results demonstrate that STM is more effective or comparable to the other models no matther with or without context. In addition, our proposed metrics outperforms others by being able to better extracing gist information and leveraging context information.

\vspace{-0.5cm}
\section{Related Work}
\vspace{-0.1cm}
%meeting summarization
Speech and dialogue summarization has become important in recent years as the number of multimedia resources containing speech has grown.
A primary goal for most speech summarization systems is to account for the special characteristics of dialogue. Early work in this area investigated supervised learning methods, including maximum entropy, conditional random fields (CRFs), and support vector machines (SVMs)~\cite{Buist04automaticsummarization,Galley:2006:SCR:1610075.1610126,Xie08evaluatingthe}. 
%  Other research focues on developing methods for selecting relevant utterances to form the summary by investigating sentence similarity measures appropriate for the speech context (e.g., ~\newcite{Gurevych:2004:SSA:1220355.1220465}). 
For unsupervised methods, maximal marginal relevance (MMR) is investigated in~\cite{Zechner:2002:ASO:638178.638181} and~\cite{Xie:2010:UCN:1857999.1858005}.~\newcite{Gillick:2009:GOF:1582709.1583197} introduce a concept-based global optimization framework by using integer linear programming (ILP). %In addition, ClusterRank~\cite{Garg_clusterrank:a} works as a modified graph-based method to take into account the conversational speech style in meetings.

%unsupervised
%For unsupervised methods,~\newcite{Zechner:2002:ASO:638178.638181} investigates speech summarization based on maximal marginal relevance (MMR) and cross-speaker linking of information.~\newcite{Lin:2010:RMF:1858681.1858690} combine supervised and unsupervised models, and formulate the extractive summarization as a risk minimization problem. In~\cite{Xie:2010:UCN:1857999.1858005}, the confusion networks are employed in an unsupervised MMR framework for speech summarization.~\newcite{Gillick:2009:GOF:1582709.1583197} introduce a concept-based global optimization framework by using integer linear programming(ILP), where concepts were used as the minimum units, and the important sentences were extracted to cover as many concepts as possible. In addition, ClusterRank~\cite{Garg_clusterrank:a} works as a modified graph-based method to take into account the conversational speech style in meetings.

Only in very recent works has decision summarization been addressed in~\cite{Fernandez},~\cite{Bui:2009:EDM:1708376.1708410} and~\cite{wang-cardie:2011}.~\cite{Fernandez} and~\cite{Bui:2009:EDM:1708376.1708410} utilize semantic parser to identify candidate phrases for decision summaries and employ SVM to rank those phrases. They also train HMM and SVM directly on a set of decision-related dialogue acts on token level and use the classifiers to identify summary-worthy words.~\newcite{wang-cardie:2011} provide an exploration on supervised and unsupervised learning for decision summarization on both utterance- and token- level.

%topic model and summarization
Our work also arises out of applying topic models to text summarization~\cite{Bhandari_Ito_Shimbo_Matsumoto_2008,Haghighi:2009:ECM:1620754.1620807,Celikyilmaz:2010:HHM:1858681.1858765,Celikyilmaz:2010:HHM:1858681.1858765}. Mostly, the sentences are ranked according to importance based on latent topic structures, and top ones are selected as the summary. %~\newcite{Wang:2009:MSU:1667583.1667675} propose a sentence-based topic model by making use of both the term-document and term-sentence assoiciations. ~\newcite{Celikyilmaz:2010:HHM:1858681.1858765} combine a hierarchical topic model and a regression model to learn the importance of each sentence.
There are some works for applying document-level topic models to speech summarization~\cite{KONG2006,Chen:2008:ESD:1332129.1332195,DBLP:conf/interspeech/Hazen11}. Different from their work, we further investigate the topic models of fine granularity on sentence level and leverage context information for decision summarization task. %In addtion, We also propose a token-level summarization framework to compare with dialogue act-level framework.

%However, most of researches focus on sentence scoring and extraction; little work has been done on token-level summarization to date. 
%topic model and meeting/speech
%~\newcite{Purver:2006:UTM:1220175.1220178} uses unsupervised topic modeling to address the problems of topic segmentation and topic identification in multi-party dialogue.

%NEW CHANGE: keyphrase extraction in meeting summarization
Most existing approaches for speech summarization result in a selection of utterances from the dialogue, which cannot remove the redundancy within utterances. To eliminate the superfluous words, our work is also inspired by keyphrase extraction of meetings~\cite{Liu:2009:UAA:1620754.1620845,DBLP:journals/taslp/LiuLL11} and keyphrase based summarization~\cite{Riedhammer:2010:LSS:1837521.1837625}. However, a small set of keyphrases are not enough to concretely display the content. Instead of only picking up keyphrases, our work identifies all of the summary-worthy words and phrases, and removes redundancies within utterances.

\vspace{-0.1cm}
\section{Summarization Frameworks}
\vspace{-0.1cm}
In this section,  we first present our proposed token-level decision summarization framework --- {\bf DomSum} --- which utilizes latent topic structure in utterances to extract words from {\bf Dom}inant Topic (see details in Section 3.1) to form {\bf Sum}maries. In Section 3.2, we describe four existing sentence scoring metrics denoted as {\it OneTopic, MultiTopic, TMMSum} and {\it KLSum} which are also based on latent topic distributions. We adopt them to the utterance-level summarization for comparison in Section 6.

%In the AMI meeting corpus~\cite{Carletta05theami}, several decisions will be made per meeting. Each decision is linked with a cluster of decision-related dialogue acts (DRDAs) that support this decision, which is called {\bf True Clusterings}. Then every cluster of DRDAs makes up one decision document. We might not know which decision(s) each DRDA supports, but we can use clustering algorithms, such as the hierachical agglomerative clustering in~\cite{wang-cardie:2011}, to generate {\bf System Clusterings}. Then one decision is created per cluster. We will test our method in both situations where the true clusterings are known and unknown.
\vspace{-0.1cm}
\subsection{Token-level Summarization Framework}
\vspace{-0.1cm}
{\bf Domsum} takes as input the clusters of DRDAs (with or without additional context DAs), the topic distribution for each DA and the word distribution for each topic. The output is a set of topic-coherent summary-worthy words which can be used directly as the summary or to further generate abstractive summary.
%{\bf DomSum} targets at selecting tokens for each DA concentrating on one topic by labeling each word as belonging to the most probable topic (denoted as ``{\it Dominant Topic}"). %This makes it both capture the crux in each DA and leverage context well to boost the summary quality according to semantic similarity.
We introduce DomSum in two steps according to its input: taking clusters of DRDAs as the input and with additional context information.

\paragraph{DRDAs Only.}
Given clusters of DRDAs, we use Algorithm~\ref{alg:domsum_DRDA} to produce the token-level summary for each cluster. Generally, Algorithm 1 chooses the topic with the highest probability as the {\it dominant topic} given the dialogue act (DA). Then it collects the words with a high joint probability with the dominant topic from that DA.

%For LocalLDA and STM, the probability of $P(T_{j}|DA_{i})$ is obtained directly from the inference results. For MG-LDA, $P(T_{j}|DA_{i})$ is computed as the expectation of local topic distributions with respect to the window distribution.
\vspace{-0.2cm}
\begin{algorithm}
\scriptsize
\SetKwFunction{Union}{Union}
\SetKwInOut{Input}{Input}
\SetKwInOut{Output}{Output}
%\linesnumbered
\Input{Cluster $C=\{DA_{i}\}$, $P(T_{j}|DA_{i})$, $P(w_{k}|T_{j})$}
\Output{Summary}
\BlankLine
Summary$\leftarrow \Phi$ (empty set)

\ForEach{$DA_{i}$ in $C$}{
DomTopic${\leftarrow \max_{T_{j}} P(T_{j}|DA_{i})}$ (*)

Candidate$\leftarrow \Phi$

\ForEach{{\normalfont word} $w_{k}$ in $DA_{i}$}{
SampleTopic${\leftarrow \max_{T_{j}} P(w_{k}|T_{j})P(T_{j}|DA_{i})}$

\If{{\normalfont DomTopic == SampleTopic}}
{Candidate $\leftarrow$ \Union{{\normalfont Candidate}, $w_{k}$}}

}
Summary $\leftarrow$ \Union{{\normalfont Summary, Candidate}}

}

\caption{\small {\bf DomSum} --- The token-level summarization framework. DomSum takes as input the clusters of DRDAs and related probability distributions.}
\label{alg:domsum_DRDA}
\end{algorithm}

\vspace{-0.5cm}
\paragraph{Leveraging Context.}
%add: how to chooose the context, need to name the center DA and context DAs
%adj/tfidf criteria
%maj/all criteria

For each DRDA (denoted as ``{\it center DA}"), we study two types of context information (denoted as ``{\it context DAs}"). One is adjacent DAs, i.e., immediately preceding and succeeding DAs, the other is the DAs having top TF-IDF similarities with the center DA. Context DAs are added into the cluster the corresponding center DA in.

We also study two criteria of word selection from the context DAs. %As the {\it dominant topic} in each context DA might not be the same as the dominant topic in corresponding center DA, 
For each context DA, we can take the words appearing in the dominant topic of either this context DA or its center DRDA. We will show in Section 6.1 that the latter performs better as it produces more topic-coherent summaries. Algorithm~\ref{alg:domsum_DRDA} can be easily modified to leverage context DAs by updating the input clusters and assigning the proper dominant topic for each DA accordingly --- this changes the step $(*)$ in Algorithm~\ref{alg:domsum_DRDA}.

%\begin{algorithm}
%\scriptsize
%\SetKwFunction{Union}{Union}
%\SetKwInOut{Input}{Input}
%\SetKwInOut{Output}{Output}
%%\linesnumbered
%\Input{Cluster $C=\{DA_{i}\}$, $P(T_{j}|DA_{i})$, $P(w_{k}|T_{j})$}
%\Output{Summary}
%\BlankLine
%Summary$\leftarrow \Phi$
%
%\ForEach{$DA_{i}$ in $C$}{
%centerDA${\leftarrow DA_{i}}$
%
%DomTopic${\leftarrow \max_{T_{j}} P(T_{j}|DA_{i})}$
%
%Candidate$\leftarrow \Phi$
%
%\ForEach{{\normalfont word} $w_{k}$ in $DA_{i}$}{
%SampleTopic${\leftarrow \max_{T_{j}} P(w_{k}|T_{j})P(T_{j}|DA_{i})}$
%
%\If{{\normalfont DomTopic == SampleTopic}}
%{Candidate $\leftarrow$ \Union{{\normalfont Candidate}, $w_{k}$}}
%
%}
%Summary $\leftarrow$ \Union{{\normalfont Summary, Candidate}}
%
%\ForEach{$DA_{con}$ in {\tt context}{\normalfont (}$DA_{i}${\normalfont )}}{
%TargetTopic${\leftarrow}$ {\tt SelectTopic}{\normalfont (}$DA_{con}${\normalfont )}\\
%\tcp{\scriptsize dominant topic of centerDA or this $DA_{con}$}
%Candidate$\leftarrow \Phi$
%
%\ForEach{{\normalfont word} $w_{k}$ in $DA_{con}$}{
%SampleTopic${\leftarrow \max_{T_{j}} P(w_{k}|T_{j})P(T_{j}|DA_{i})}$
%
%\If{{\normalfont TargetTopic == SampleTopic}}
%{Candidate $\leftarrow$ \Union{{\normalfont Candidate}, $w_{k}$}}
%
%}
%
%}
%Summary $\leftarrow$ \Union{{\normalfont Summary, Candidate}}
%}
%
%\caption{\small Leveraging Context for {\bf DomSum}. The input includes a set of clusters consisting of decision-related dialogue acts (DRDAs) and corresponding context DAs, and related probability distributions.}
%\label{alg:domsum_context}
%\end{algorithm}

\subsection{Utterance-level Summarization Metrics}
We also adopt four sentence scoring metrics based on the latent topic structure for extractive summarization. Though they are developed on different topic models, given the desired topic distributions as input, they can rank the utterances according to their importance and provide utterance-level summaries for comparison.
%There exist various sentence scoring metrics based on the latent topic structure of documents for extractive summarization~\cite{Bhandari_Ito_Shimbo_Matsumoto_2008,Chen:2008:ESD:1332129.1332195,Haghighi:2009:ECM:1620754.1620807}. Though they are based on different topic models, we can adopt their metrics for ranking utterances according to their importance in spoken meeting and extract the top ones as the summary.

\paragraph{OneTopic and MultiTopic.}
In~\cite{Bhandari_Ito_Shimbo_Matsumoto_2008}, several sentence scoring functions are introduced based on Probabilistic Latent Semantic Indexing. We adopt two metrics, which are {\it OneTopic} and {\it MultiTopic}.
%The {\it OneTopic} and {\it MultiTopic} metrics are introduced in~\cite{Bhandari_Ito_Shimbo_Matsumoto_2008} based on Probabilistic Latent Semantic Indexing (PLSI). %Though both of them are for text summarization, we accomodate the two sentence scoring metrics to decision summarization. 
For OneTopic, topic $T$ with highest probability $P(T)$ is picked as the central topic per cluster $C$. The score for $DA$ in $C$ is: %computed as:

\vspace{-0.4cm}
{\scriptsize
\begin{align*}
P(DA|T)=\frac{\sum_{w\in DA}P(T|DA,w)}{\sum_{DA'\in C, w\in DA'}P(T|DA',w)},
\end{align*}
}

\vspace{-0.4cm}
MultiTopic modifies OneTopic by taking all of the topics into consideration. Given a cluster $C$, $DA$ in $C$ is scored as:

\vspace{-0.4cm}
{\scriptsize
\begin{align*}
\sum_{T}P(DA|T)P(T)=\sum_{T}\frac{\sum_{w\in DA}P(T|DA,w)}{\sum_{DA'\in C, w\in DA'}P(T|DA',w)}P(T)
\end{align*}
}

\vspace{-0.4cm}
\paragraph{TMMSum.}
\newcite{Chen:2008:ESD:1332129.1332195} propose a Topical Mixture Model (TMM) for speech summarization, where each dialogue act is modeled as a TMM for generating the document. TMM is shown to provide better utterance-level extractive summaries for spoken documents than other conventional unsupervised approaches, such as Vector Space Model (VSM)~\cite{Gong:2001:GTS:383952.383955}, Latent Semantic Analysis (LSA)~\cite{Gong:2001:GTS:383952.383955} and Maximum Marginal Relevance (MMR)~\cite{Murray05extractivesummarization}. The importance of a sentence $S$ can be measured by its generative probability $P(D|S)$, where $D$ is the document $S$ belongs to. %We will compare against their system directly and also adapt the sentence scoring function to the fine-grained topic models. 
In our experiments, one decision is made per cluster of DAs. So we adopt their scoring metric to compute the generative probability of the cluster $C$ for each $DA$:

\vspace{-0.4cm}
{\scriptsize
\begin{align*}
P(C|DA)=\prod_{w_{i}\in C} \sum_{T_{j}} P(w_{i}|T_{j})P(T_{j}|DA),
\end{align*}
}

\vspace{-0.4cm}
\paragraph{KLSum.}
Kullback-Lieber (KL) divergence is explored for summarization in~\cite{Haghighi:2009:ECM:1620754.1620807} and~\cite{Lin2010}, where it is used to measure the distance of distributions between the document and the summary. %\newcite{Haghighi:2009:ECM:1620754.1620807} show that the latent topic distribution can better reflect the content of documents than the raw unigram distribution when using the KL-divergence. 
For a cluster $C$ of DAs, given a length limit $\theta$, a set of DAs $S$ is selected as:

\vspace{-0.5cm}
{\scriptsize
\begin{align*}
S^{*}=\argmin_{S:|S|<\theta}KL(P_{C}||P_{S})=\argmin_{S:|S|<\theta}\sum_{T_{i}} P(T_{i}|C)log\frac{P(T_{i}|C)}{P(T_{i}|S)}
\end{align*}
}
%\paragraph{General Framework --- Dialogue Act-Level}
%In implementation, {\it OneTopic, MultiTopic} and {\it TMMSum} are used for ranking dialogue acts and extracting the top ones to form summaries. For {\it KLSum}, globally optimizing the criterion is exponential in the total number of DAs, we opt instead for a simple approximation where DAs are greedily added to a summary so long as they decrease KL-divergence. In order to accommodate those metrics to our task, we apply them by using the general framework in Algorithm 3. The input, achieved from topic modeling, includes clusters of DAs, the topic distribution for each DA and the word distribution for each topic. The number of DAs selected, $\theta$, is pre-specified.
%
%\begin{algorithm}
%\scriptsize
%\SetKwFunction{Union}{Union}
%\SetKwInOut{Input}{Input}
%\SetKwInOut{Output}{Output}
%
%\Input{Cluster $C=\{DA_{i}\}$, $P(T_{j}|DA_{i})$, $P(w_{i}|T_{j})$, $\delta$, $metric\in$ {$\{OneTopic, MultiTopic, TMMSum, KLSum\}$}}
%\Output{Summary}
%\BlankLine
%Summary$\leftarrow \Phi$
%
%\ForEach{$DA_{i}$ in $C$}{
%ComputeScore($DA_{i}$, $metric$)
%}
%Rank DAs By Score;
%
%\For{$i\leftarrow 1$ \KwTo $\delta$}{
%  \eIf{$metric == KLSum$}
%  {Candidate $\leftarrow$ SelectDAByGreedyAlgorithm}
%  {Candidate $\leftarrow$ SelectDARankAt $i$}
%  
%  Summary $\leftarrow$ \Union{{\normalfont Summary, Candidate}}
%}
%
%\caption{\small A general framework used for integrating the dialogue act (DA) scoring metrics (OneTopic, MultiTopic, TMMSum and KLSum) into decision summarization.}
%\label{alg_general}
%\end{algorithm}
\vspace{-0.3cm}
\section{Topic Models}
\vspace{-0.1cm}
In this section, we briefly describe the three fine-grained topic models employed to compute the latent topic distributions on utterance level in the meetings. %They are all based on Latent Dirichlet Allocation (LDA)~\cite{Blei:2003:LDA:944919.944937}, which is a probabilistic generative model representing the documents as the mixtures over latent topics. %LDA can model the word co-occurence at the meeting level, but as we will show, the decision topics are more likely to be discovered at the utterance level. %The three models described below will exploit word co-occurrence on the utterance level.%   DAs rather than the whole meetings, which can distill the salient topics for each DA.
According to the input of Algorithm~\ref{alg:domsum_DRDA}, we are interested in estimating the topic distribution for each DA $P(T|DA)$ and the word distribution for each topic $P(w|T)$. %For LocalLDA and STM, $P(T|DA)$ is obtained directly from the inference results. 
For MG-LDA, $P(T|DA)$ is computed as the expectation of local topic distributions with respect to the window distribution.

\vspace{-0.1cm}
\subsection{Local LDA}
\vspace{-0.1cm}
Local LDA (LocalLDA)~\cite{Brody:2010:UAM:1857999.1858121} uses almost the same probabilistic generative model as Latent Dirichlet Allocation (LDA)~\cite{Blei:2003:LDA:944919.944937}, except that it treats each sentence as a separate document\footnote{For the generative process of LDA, the DAs in the same meeting make up the document, so ``each DA" is changed to ``each meeting" in LocalLDA's generative process.}. Each DA $d$ is generated as follows:

\vspace{-0.2cm}
\begin{enum}
  \item For each topic $k$:
  \begin{enum}
  \item Choose word distribution: $\phi_{k}\sim Dir(\beta)$
  \end{enum}
  \item For each DA $d$:
  \begin{enum}
  \item Choose topic distribution: $\theta_{d}\sim Dir(\alpha)$
  \item For each word $w$ in DA $d$:
  \begin{enum}
    \item Choose topic: $z_{d,w}\sim \theta_{d}$
      \item choose word: $w\sim\phi_{z_{d,w}}$
  \end{enum}
    
  \end{enum}
\end{enum}

\vspace{-0.4cm}
\subsection{Multi-grain LDA}
\vspace{-0.1cm}
Multi-grain LDA (MG-LDA)~\cite{Titov:2008:MOR:1367497.1367513} can model both the meeting specific topics (e.g. the design of a remote control) and various concrete aspects (e.g. the cost or the functionality). The generative process is:

\vspace{-0.2cm}
\begin{enum}
  \item Choose a global topic distribution: $\theta_{m}^{gl}\sim Dir(\alpha^{gl})$
  \item For each sliding window $v$ of size $T$:
  \begin{enum}
    \item Choose local topic distribution: $\theta_{m,v}^{loc}\sim Dir(\alpha^{loc})$
    \item Choose granularity mixture: $\pi_{m,v}\sim Beta(\alpha^{mix})$
  \end{enum}
  
  \item For each DA $d$:
  \begin{enum}
      \item choose window distribution: $\psi_{m,d}\sim Dir(\gamma)$  
  \end{enum}  

  \item For each word $w$ in DA $d$ of meeting $m$:
  \begin{enum}
    \item Choose sliding window: $v_{m,w}\sim \psi_{m,d}$
    \item Choose granularity: $r_{m,w}\sim \pi_{m,v_{m,w}}$
    \item If $r_{m,w}=gl$, choose global topic: $z_{m,w}\sim \theta_{m}^{gl}$
    \item If $r_{m,w}=loc$, choose local topic: $z_{m,w}\sim \theta_{m,v_{m,w}}^{loc}$
    \item Choose word $w$ from the word distribution: $\phi_{z_{m,w}}^{r_{m,w}}$
  \end{enum}
\end{enum}

\vspace{-0.4cm}
\subsection{Segmented Topic Model}
\vspace{-0.1cm}
The last model we utilize is Segmented Topic Model (STM)~\cite{Du:2010:STM:1842816.1842821}, which jointly models document- and sentence-level latent topics using a two-parameter Poisson Dirichlet Process (PDP). %By using STM, we can take into account the segments and corresponding layout in the meetings which provides structural information. 
Given parameters $\alpha, \gamma, \Phi$ and PDP parameters $a,b$, the generative process is:

\vspace{-0.2cm}
\begin{enum}
  \item Choose distribution of topics: $\theta_{m}\sim Dir(\alpha)$
  \item For each dialogue act $d$:
     \begin{enum}
       \item Choose distribution of topics: $\theta_{d}\sim PDP(\theta_{m},a,b)$
     \end{enum}
  \item For each word $w$ in dialogue act $d$:
     \begin{enum}
       \item Choose topic: $z_{m,w}\sim \theta_{d}$
       \item Choose word: $w\sim \phi_{z_{m,w}}$
     \end{enum}
     
\end{enum}

\vspace{-0.4cm}
\section{Experimental Setup}
\vspace{-0.1cm}
\paragraph{The Corpus.}  We evaluate our approach on the AMI meeting
corpus~\cite{Carletta05theami} that consists of 140 multi-party
meetings. % with a wide range of annotations. 
The 129 scenario-driven meetings involve four participants playing different roles on a design team. 
A short (usually one-sentence) abstract is manually constructed to summarize each decision discussed in the meeting and used as gold-standard summaries in our experiments.
%Importantly, the corpus includes a short (usually one-sentence),
%manually constructed abstract summarizing each decision discussed in
%the meeting.  In addition, all of the dialogue acts that support
%(i.e., are relevant to) each decision are annotated as such. We use
%the manually constructed decision abstracts as gold-standard
%summaries.

\vspace{-0.1cm}
\paragraph{System Inputs.} Our summarization
system requires as input a partitioning of the DRDAs according to the
decision(s) that each supports (i.e., one cluster of DRDAs per
decision). As mentioned earlier, we assume for all experiments that
the DRDAs for each meeting have been
identified.  For evaluation we consider two system input settings.
In the {\bf True Clusterings} setting, we use the AMI annotations to
create perfect partitionings of the DRDAs as the input; in the {\bf System Clusterings} setting, we
employ a hierarchical agglomerative clustering algorithm used for this
task in previous work \cite{wang-cardie:2011}.
The~\newcite{wang-cardie:2011} clustering method
groups DRDAs according to their LDA topic distribution similarity.  As
better approaches for DRDA clustering become available, they could be
employed instead.

\vspace{-0.1cm}
\paragraph{Evaluation Metric.} To evaluate the performance of various summarization approaches, we use the widely accepted ROUGE~\cite{Lin:2003:AES:1073445.1073465} metrics. %which is shown to correlate with human rankings. ROUGE estimates the coverage of content in the system summary by comparing it to the gold-standard summary and can identify systems producing succinct and descriptive summaries. 
We use the stemming
option of the ROUGE software at \url{http://berouge.com/} and remove
stopwords from both the system and gold-standard summaries, same as~\newcite{Riedhammer:2010:LSS:1837521.1837625} do.
%ROUGE-1, ROUGE-2 and ROUGE-SU4 are considered in experiments.
%ROUGE-1 (unigram overlap), ROUGE-2 (bigram overlap) and ROUGE-SU4 (unigram overlap and skip bigram overlap where bigrams can consist of non-contiguous words intervened by at most four words) are considered.

\vspace{-0.1cm}
\paragraph{Inference and Hyperparameters}
We use the implementation from~\cite{OttLu} for the three topic models in Section 4. The collapsed Gibbs Sampling approach~\cite{griffiths_steyvers04} is exploited for inference. %According to~\cite{Haghighi:2009:ECM:1620754.1620807}, for a more specific distribution, its concentration parameter should be set comparatively smaller. 
Hyperparameters are chosen according to~\cite{Brody:2010:UAM:1857999.1858121},~\cite{Titov:2008:MOR:1367497.1367513} and~\cite{Du:2010:STM:1842816.1842821}. In LDA and LocalLDA, $\alpha$ and $\beta$ are both set to $0.1$ . For MG-LDA, $\alpha^{gl}$, $\alpha^{loc}$ and $\alpha^{mix}$ are set to $0.1$; $\gamma$ is 0.1 and the window size $T$ is 3. And the number of local topic is set as the same number of global topic as discussed in~\cite{Titov:2008:MOR:1367497.1367513}. In STM, $\alpha$, $a$ and $b$ are set to $0.5$, $0.1$ and $1$, respectively.

\vspace{-0.1cm}
\subsection{Baselines and Comparisons}
\vspace{-0.2cm}
%A series of comparison experiments are conducted for three purposes. (1) compare the fine-grained topic models and the models discovering latent structure on document (or meeting) level (2) compare the token-level summarization metrics and the utterance-level summarization metrics (3) 

%Besides investigating the context information, 
We compare our token-level summarization framework based on the fine-grained topic models to (1) two unsupervised
baselines, (2) token-level summarization by LDA, (3) utterance-level summarization by Topical Mixture Model (TMM)~\cite{Chen:2008:ESD:1332129.1332195}, 
(4) utterance-level summarization based on the fine-grained topic models using existing metrics (Section 3.2), (5) two supervised methods, and (6) an
upperbound derived from the AMI gold standard decision
abstracts. (1) and (6) are described below, others will be discussed in Section 6.

\vspace{-0.1cm}
\paragraph{The {\sc Longest DA} Baseline.} As in~\cite{Riedhammer:2010:LSS:1837521.1837625} and~\cite{wang-cardie:2011},
this baseline simply selects the longest DRDA in each cluster as the
summary. Thus, it performs utterance-level decision
summarization. %Though it's possible that decision content is spread over multiple DRDAs in the cluster, 
This baseline and the next allow
us to determine summary quality when summaries are restricted to a
single utterance.

\vspace{-0.1cm}
\paragraph{The {\sc Prototype DA} Baseline.} Following \newcite{wang-cardie:2011},
the second baseline selects the decision cluster prototype (i.e., the
DRDA with the largest TF-IDF similarity with the cluster centroid) as
the summary. %Like the first baseline, it performs utterance-level decision summarization.

%\paragraph{Latent Dirichlet Allocation (LDA)}
%The three fine-grained topic models are all based on Latent Dirichlet Allocation (LDA)~\cite{Blei:2003:LDA:944919.944937}, which models the topic structures on document level. We use our token-level summarization metrics (DomSum) on LDA in the comparison.

%\paragraph{Topical Mixture Model (TMM)}
%As mentioned in Section 3.2, ~\cite{Chen:2008:ESD:1332129.1332195} also models the topic structure on the utterance level and ranks the sentences by their generative score. We reimplement this approach for direct comparison.

%\paragraph{Supervised Learning (SVMs and CRFs).} For a better comparison, we also compare our approach 
%to two supervised learning methods --- Support
%Vector Machines~\cite{citeulike:3340317} with RBF kernel and order-1 Conditional Random
%Fields~\cite{Lafferty:2001:CRF:645530.655813} --- trained using the
%same features as~\cite{wang-cardie:2011} to identify the summary-worthy {\bf
%  tokens} to include in the decision abstract. %\footnote{We use the SVM with an RBF kernel and employ an order-1 CRF.} 
%Three-fold cross validation is conducted for both methods.
\vspace{-0.1cm}
\paragraph{Upperbound.} We also compute an upperbound that reflects the gap 
between the best possible extractive summaries and the human-written
abstracts according to the ROUGE score: for each cluster of DRDAs, we
select the words that also appear in the associated decision abstract.

\section{Results and Discussion}
%In this section, we will present two sets of experiments. one in which each DRDA is paired with the decision(s) it supports ({\bf True Clusterings}), and another in which an automated system does the DRDA-decision pairing ({\bf System Clusterings}). 
\vspace{-0.1cm}
\subsection{True Clusterings}
\vspace{-0.1cm}
\paragraph*{How do fine-grained topic models compare to basic topic models or baselines?}
%\subsection*{Fine-grained Models vs. Baselines/ LDA/ TMM}
%We adopt the two unsupervised baselines from \cite{wang-cardie:2011}. The first method simply returns the longest DRDA in the cluster as the summary ({\sc Longest DA}). The second approach returns the decision cluster prototype, i.e., the DRDA with the largest TF-IDF similarity with the cluster centroid ({\sc Prototype DA}). Both unsupervised methods allow us to determine summary quality when summaries are restricted to a single utterance.

\begin{figure}
\hspace{-0.5cm}
\includegraphics[width=3.3in, height=2.15in]{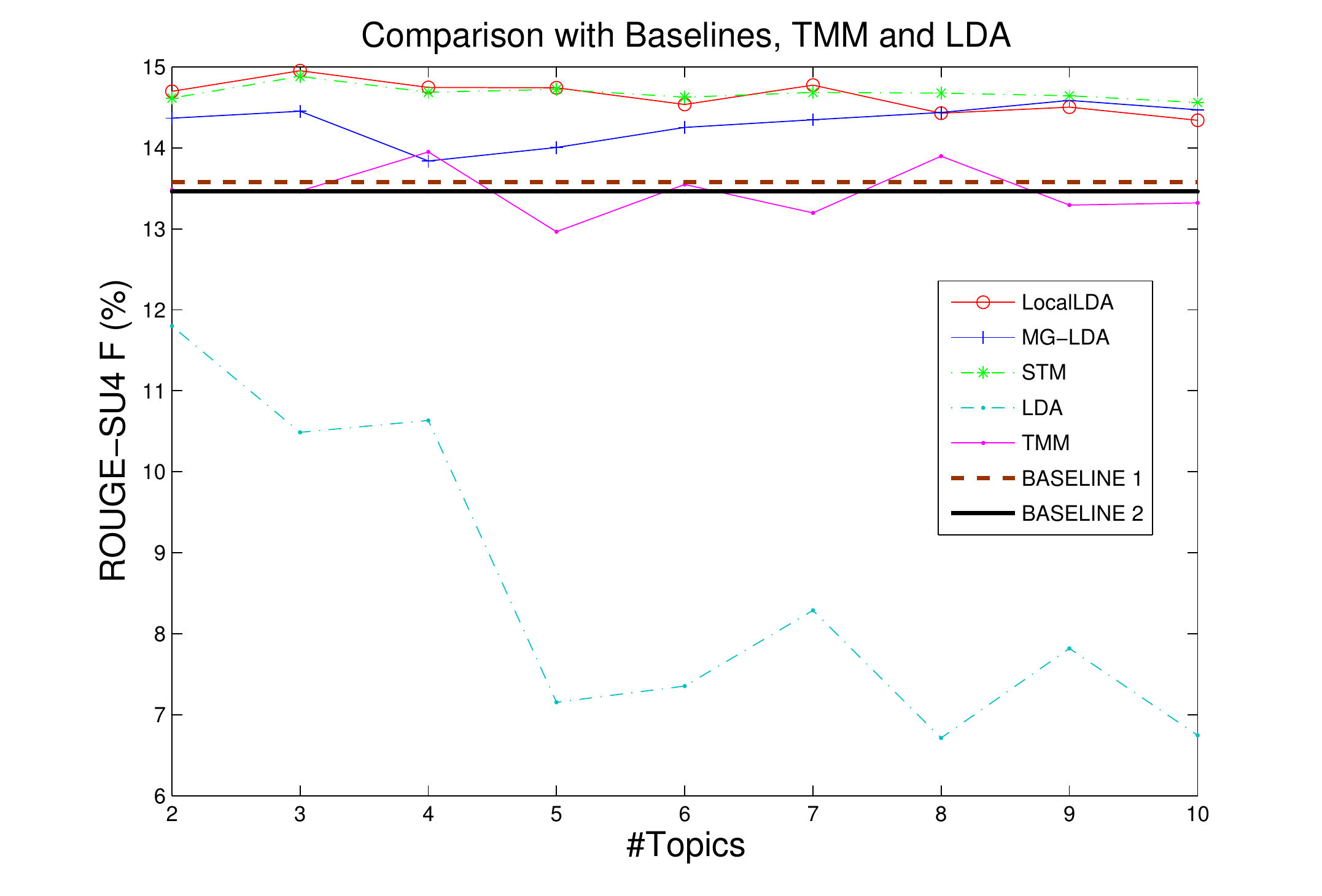}
\vspace{-0.7cm}
\caption{\footnotesize With true clusterings of DRDAs as the input, we use DomSum to compare the performance of LocalLDA, MGLDA and STM against two baselines, LDA and TMM. ``{\# topic}" indicates the number of topics for the model. For MGLDA, ``{\# topic}" is the number of local topics.}
\label{fig:Exp1}
\end{figure}

Figure~\ref{fig:Exp1} demonstrates that by using the DomSum token-level summarization framework, the three fine-grained topic models uniformly outperform the two non-trivial baselines and TMM~\cite{Chen:2008:ESD:1332129.1332195} (reimplemented by us) that generates utterance-level summaries. Moreover, the fine-grained models also beat basic LDA under the same DomSum token-level summarization framework. This shows the fine-grained topic models that discover topic structures on utterance-level better identify gist information. %Notice also that STM achieves better results than LocalLDA or MG-LDA, and has a relative stable performance when varying the number of topics. 

\begin{figure}
\hspace{-0.5cm}
\includegraphics[width=3.3in, height=2.15in]{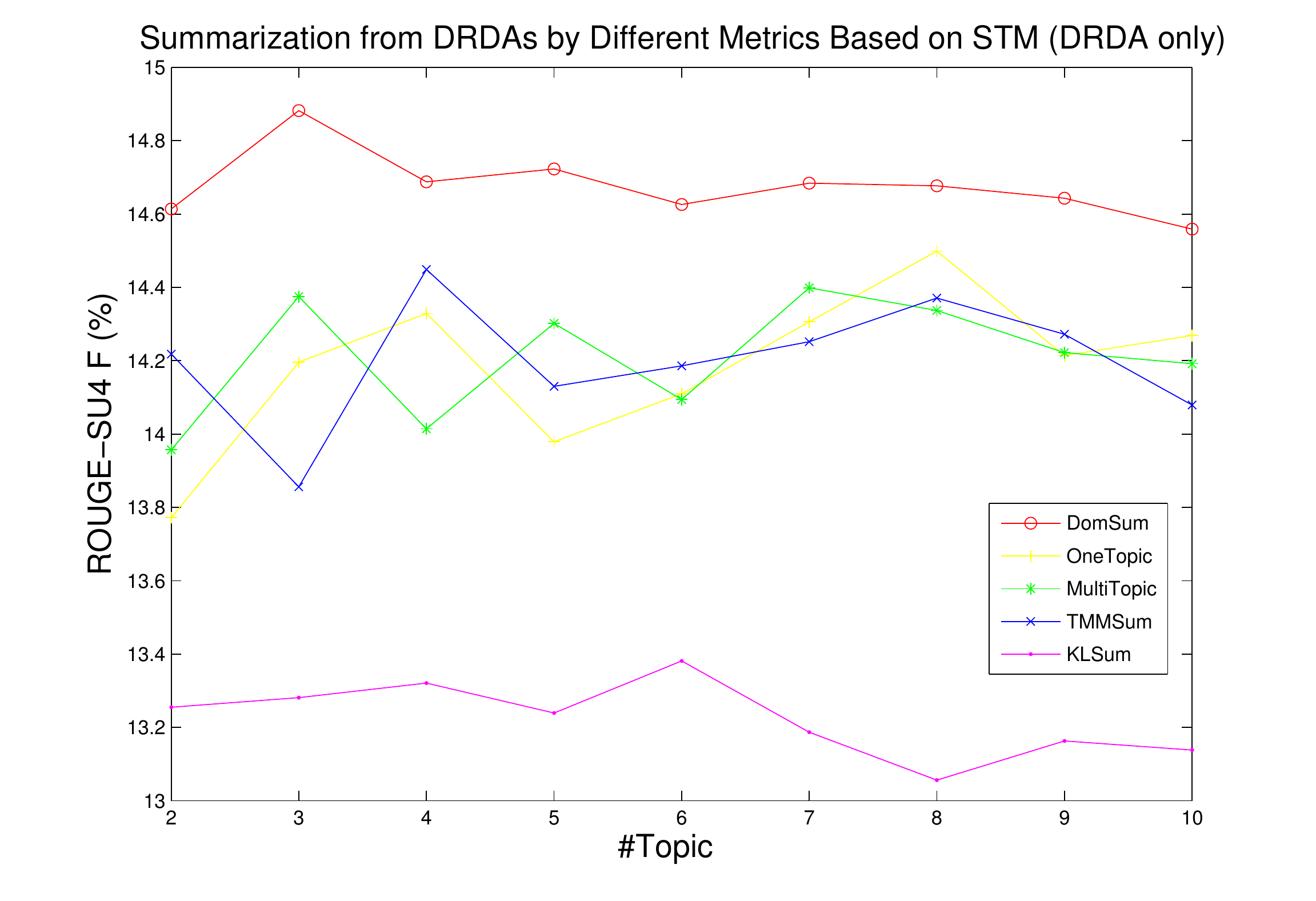}
\vspace{-0.7cm}
\caption{\footnotesize With true clusterings of DRDAs as the input, DomSum is compared with four DA-level summarization metrics using topic distributions from STM. Results from LocalLDA and MGLDA are similar so they are not displayed.}
\label{fig:Exp2}
\end{figure}

%\subsection*{Fine-grained Models vs. LDA }
%We compare the three fine-grained topic models with LDA which models topic structure on meeting level. Figure 2 illustrates that fine-grained topic models outperforms LDA. This confirms that discovering topics in spoken meeting at finer granulairty benefit the summarization process.
%\begin{figure}
%\includegraphics[width=3in, height=2in]{figs/DomNon.eps}
%\caption{With true clusterings of DRDAs as the input, we use DomSum compare the performance of localLDA, MGLDA and STM against LDA. ``{\# topic}" indicates the number of topics for the model. For MGLDA, ``{\# topic}" is the number of local topics.}
%\label{fig:TMAna}
%\end{figure}
\vspace{-0.1cm}
\paragraph*{Can the proposed token-level summarization framework better identify important words and remove redundancies than utterance selection methods?}
%\subsection*{Token-level vs. DA-level Summarization Metrics}
Figure~\ref{fig:Exp2} demonstrates the comparison results for our DomSum token-level summarization framework with four existing utterance scoring metrics discussed in Section 3.2, namely OneTopic, MultiTopic, TMMSum and KLSum. The utterance with highest score is extracted to form the summary. LocalLDA and STM are utilized to compute the input distributions, i.e., $P(T|DA)$ and $P(w|T)$. From Figure~\ref{fig:Exp2}, DomSum yields the best F scores which shows that the token-level summarization approach is more effective than utterance-level methods.

\begin{figure}
\includegraphics[width=3.3in, height=2.15in]{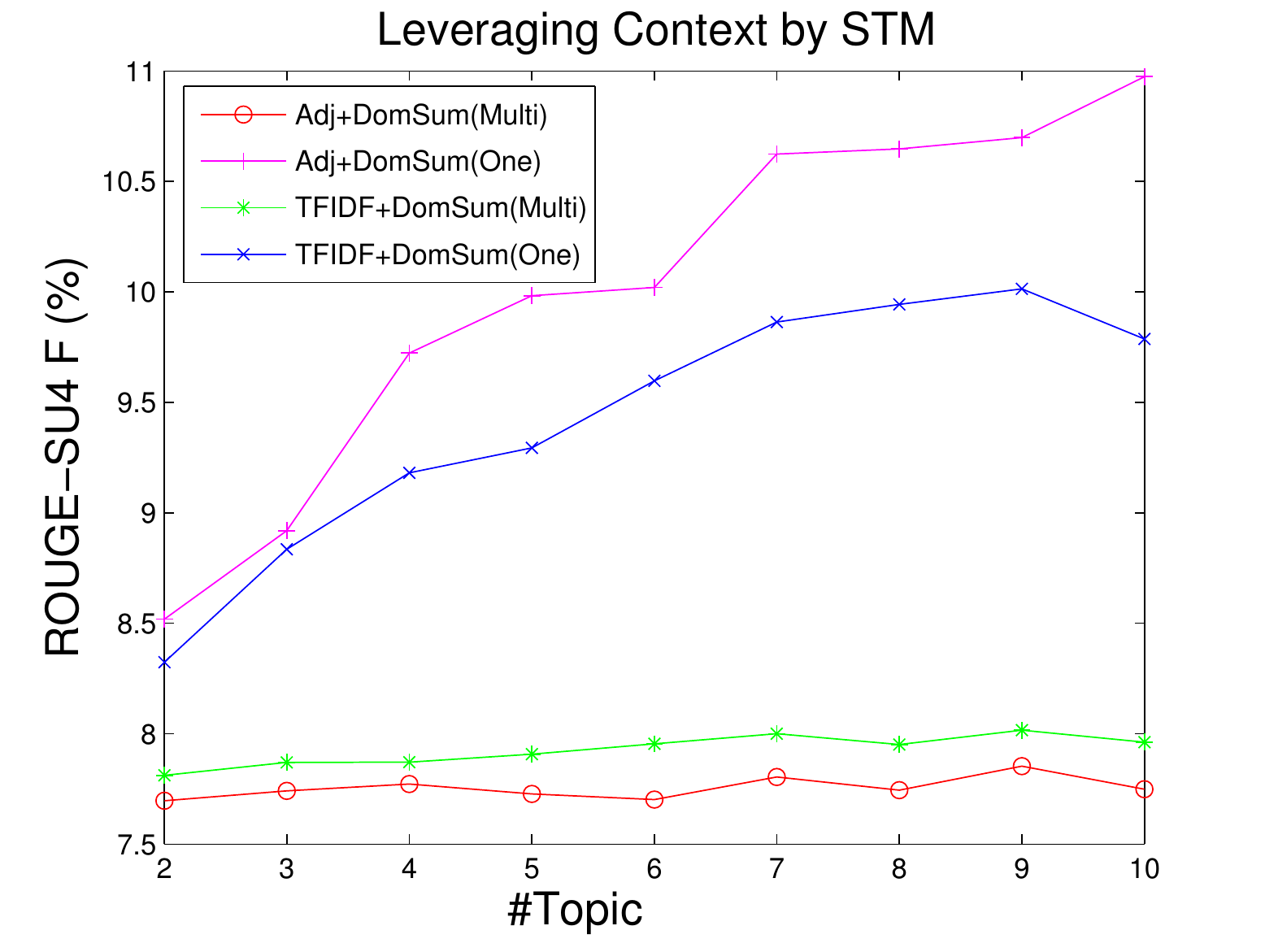}
\vspace{-0.7cm}
\caption{\footnotesize Under DomSum framework, two types of context information are added: Adjacent DA (``Adj") and DAs with high TFIDF similarities (``TFIDF"). For each context DA, selecting words from the dominant topic of center DA (``One") or the current context DA (``Multi") are investigated.}
%\caption{\footnotesize ``Adj/TFIDF": 10 adjacent DAs as context/ 10 DAs with top TFIDF similarties as context; ``DomSum(Multi)/ DomSum(One)": for each context DA, the words labeled with the dominant topic of the context DA/ center DRDA are selected.}
\label{fig:Exp3}
\end{figure}

\begin{figure}
\vspace{-0.3cm}
\includegraphics[width=3.3in, height=2.15in]{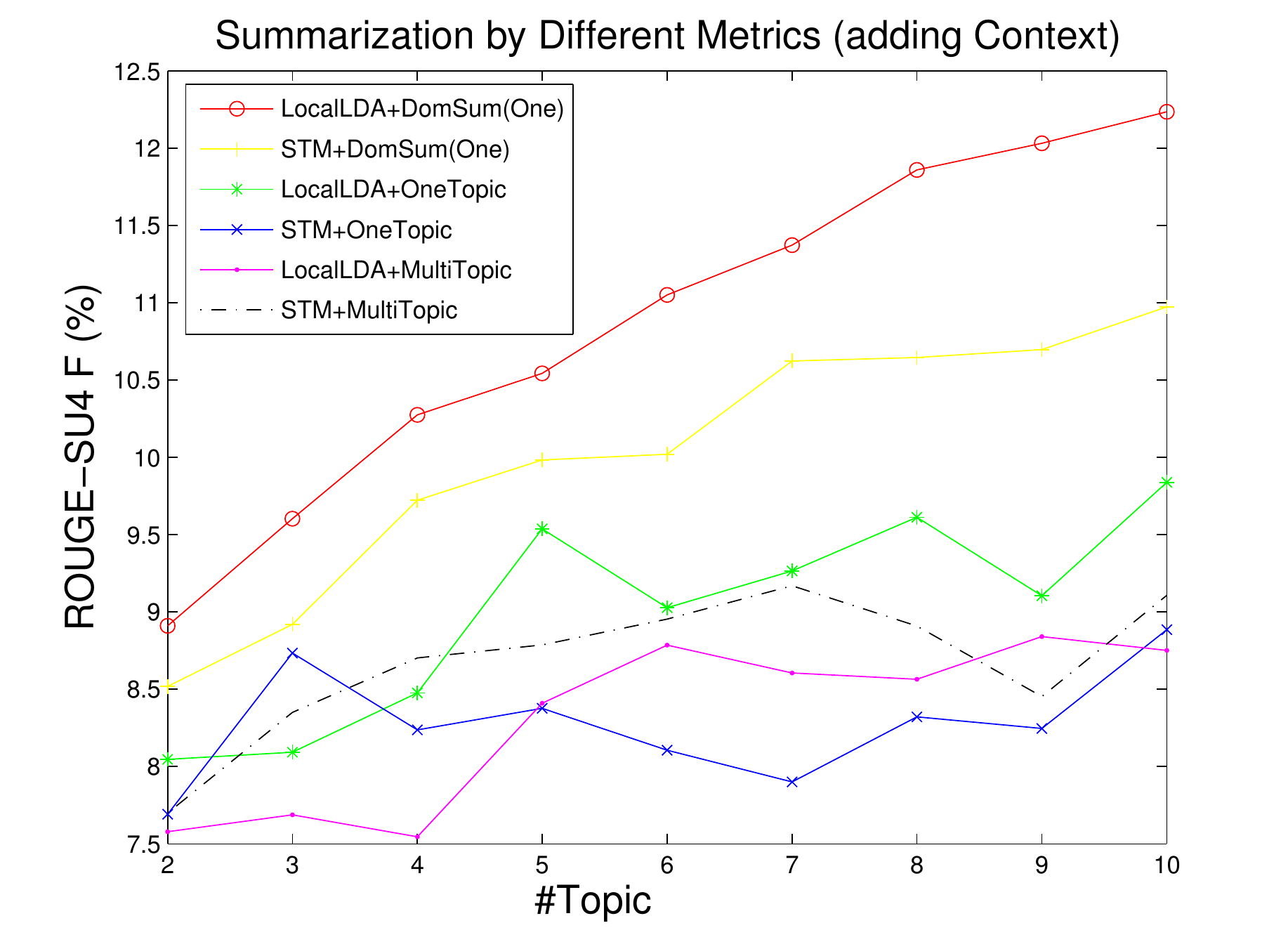}
\vspace{-0.7cm}
\caption{\footnotesize By using adjacent DAs as context, DomSum is compared with two DA-level summarization metrics: OneTopic and MultiTopic. For DomSum, the words of context DA from dominant topic of the center DA (``One") is selected; For OneTopic and MultiTopic, three top ranked DAs are selected.}
\label{fig:Exp4}
\end{figure}

\vspace{-0.1cm}
\paragraph*{Which way is better for leveraging context information?}
%\subsection*{No context vs. Leveraging Context}
We explore two types of context information. For adjacent content ({\it Adj} in Figure~\ref{fig:Exp3}), 5 DAs immediately preceding and 5 DAs succeeding the center DRDA are selected. For TF-IDF context ({\it TFIDF} in Figure~\ref{fig:Exp3}), 10 DAs of highest TF-IDF similarity with the center DRDA are taken. We also explore two ways to extract summary-worthy words from the context DA --- selecting words from the dominant topic of either the center DA (denoted as ``One" in parentheses in Figure~\ref{fig:Exp3}) or the current context DA (denoted as ``multi" in parentheses in Figure~\ref{fig:Exp3}). %The recalls for leveraging context based on STM is exhibited in Figure 4. It's evident that dialogue context is beneficial for retrieving valuable information.
Figure~\ref{fig:Exp3} indicates that the two types of context information do not have significant difference, while selecting the words from the dominant topic of the center DA results in better ROUGE-SU4 F scores.
Notice that compared with Figure~\ref{fig:Exp2}, the results in Figure~\ref{fig:Exp3} have lower F scores when using the true clusterings of DRDAs. This is because context DAs bring in relevant words as well as noisy information. We will show in Section 6.2 that when true clusterings are not available, the context information can boost both recall and F score.

%\begin{figure*}[htbp]
%\hspace{-0.9cm}
%\includegraphics[width=2.3in, height=1.6in]{figs/4comb_lda.eps}
%\includegraphics[width=2.3in, height=1.6in]{figs/4comb_mglda.eps}
%\includegraphics[width=2.3in, height=1.6in]{figs/4comb_stm.eps}
%\caption{\footnotesize ``Adj/TFIDF": 10 adjacent DAs as context/ 10 DAs with top TFIDF similarties as context; ``DomSum(Multi)/ DomSum(One)": for each context DA, the words labeled with the dominant topic of the context DA/ center DRDA are selected.}
%\label{fig:Exp3}
%\end{figure*}

%\begin{figure}
%\includegraphics[width=3in, height=2in]{figs/recallstm.eps}
%\caption{\footnotesize Comparison of the recall between with and without context.}
%\label{fig:TMrecall}

%\end{figure}
\begin{spacing}{0.9}
\begin{table}
%    \hspace{0.2cm}
    {\scriptsize
    \setlength{\baselineskip}{0pt}
    \begin{tabular}{|p{0.7in}|p{0.32in}|p{0.32in}|p{0.32in}|p{0.32in}|p{0.32in}|}
        \hline
\textbf{ }&\multicolumn{5}{|c|}{\textbf{True Clusterings}}\\
\hline
\textbf{ }&\multicolumn{3}{|c|}{\textbf{R-1}}&\textbf{R-2}&\textbf{R-SU4}\\
%\cline{3-7}
\hline
 &PREC&REC&F1&F1&F1\\
    {\bf Baselines}& & & & &\\
  Longest DA& 34.06  &31.28  &32.61&12.03 &13.58\\
  Prototype DA&40.72  &28.21  &33.32 &12.18 &13.46\\

\hline\hline
    
    {\bf Supervised Methods} & & &  & &\\
    CRF&52.89  &26.77  &35.53 &11.48 &14.03\\
    SVM&43.24  & 37.92  & 40.39 & 12.78 & 16.24\\

\hline\hline
     {\bf Our Approach} & & &  & &\\
     {\bf 5 topics} & & &  & &\\
    LocalLDA& 35.18  &38.92  &36.95 &12.33 &14.74\\
    \ \ \ \ + {\it context} &\ \ \ 17.26  &\ \ \ 45.34  &\ \ \ 25.00 &\ \ \ 8.40 &\ \ \ 11.05\\
    STM & 34.06  &41.30  &{\bf 37.32} &{\bf 12.42} &{\bf 14.82}\\
    \ \ \ \ + {\it context} &\ \ \ 15.60  &\ \ \ \textbf{\textit{48.10}}  &\ \ \ 23.56 &\ \ \ 8.16 &\ \ \ 9.98\\

\hline
    {\bf 10 topics} & & &  & &\\
    LocalLDA & 36.20  &36.81  &36.50 &12.04 &14.34\\
    \ \ \ \ + {\it context} &\ \ \ 21.82  &\ \ \ 41.57  &\ \ \ 28.62 &\ \ \ 9.61 &\ \ \ 12.24\\
    STM & 34.15  &40.83  &37.19 & 12.40&14.56\\
    \ \ \ \ + {\it context} &\ \ \ 17.87  &\ \ \ 46.57  &\ \ \ 25.82 &\ \ \ 8.89 &\ \ \ 10.97\\

\hline\hline
  {\bf Upperbound} &100.00 & \textbf{\textit{\underline{45.05}}}& 62.12& 33.27&34.89\\

\hline
    \end{tabular}
    }
    \vspace{-0.2cm}
    \caption{\footnotesize ROUGE-1 (R-1), ROUGE-2 (R-2) and ROUGE-SU4 (R-SU4) scores for our proposed token-level summarization approaches along with two baselines, supervised methods and the Upperbound (only using DRDAs). --- all use True Clusterings}
    \label{tab:rouge comparison results_true}
\end{table}
\end{spacing}

\vspace{-0.1cm}
\paragraph*{How do the token-level summarization framework compared to utterance selection methods for leveraging context?}
%\subsection*{Token-level vs. DA-level Summarization Metrics (Leveraging Context)}
%This group of experiments is conducted for measuring the ability of each metrics to leverage context. DomSum, OneTopic and MultiTopic are examined as they perform better than or comparable to the other two metrics. After adding context, 3 DAs will be picked up for OneTopic and MultiTopic. Figure~\ref{fig:Exp4} demonstrates the combination of LocalLDA and STM with each of the metrics. Obviously, DomSum, as a token-level summarization metrics, dominates other two metrics in leveraging context.
We also compare the ability of leveraging context of DomSum to utterance scoring metrics, i.e., OneTopic and MultiTopic. 5 DAs preceding and 5 DAs succeeding the center DA are added as context information. For context DA under DomSum, we select words from the dominant topic of the center DA (denoted as ``One" in parentheses in Figure~\ref{fig:Exp4}). For OneTopic and MultiTopic, the top 3 DAs are extracted as the summary. Figure~\ref{fig:Exp4} demonstrates the combination of LocalLDA and STM with each of the metrics. DomSum, as a token-level summarization metrics, dominates other two metrics in leveraging context.

\vspace{-0.3cm}
\paragraph*{How do our approach perform when compared with supervised learning approaches?}
%\subsection*{Fine-grained Models vs. Supervised Learning and Upperbound}
For a better comparison, we also provide summarization results by using supervised systems along with an upperbound. We use Support Vector Machines~\cite{citeulike:3340317} with RBF kernel and order-1 Conditional Random Fields~\cite{Lafferty:2001:CRF:645530.655813} --- trained with the same features as~\cite{wang-cardie:2011} to identify the summary-worthy {\bf tokens} to include in the abstract. %\footnote{We use the SVM with an RBF kernel and employ an order-1 CRF.}
A three-fold cross validation is conducted for both methods.
ROUGE-1, ROUGE-2 and ROUGE-SU4 scores are listed in Table~\ref{tab:rouge comparison results_true}. %Because our approach aims to identify summary-worthy words, ROUGE-1 is more proper to demonstrate the unigram overlap. 
%We also display ROUGE-1 precision and recall to illustrate each approach's ability to identify relevant words. %Notice that the comparison is conducted between the summary-worthy words extracted from spontaneous speech and written-language abstracts, little overlap is expected to show in terms of bigrams because of the intrinstic differences in styles~\cite{Riedhammer:2010:LSS:1837521.1837625}. So the ROUGE-2 and ROUGE-SU4 results in Table~\ref{tab:rouge comparison results_true} are lower than ROUGE-1. 
From Table~\ref{tab:rouge comparison results_true}, our token-level summarization approaches based on LocalLDA and STM are shown to outperform the baselines and even the CRF. 
Meanwhile, by adding context information, both LocalLDA and STM can get better ROUGE-1 recall than the supervised methods, even higher than the provided upperbound which is computed by only using DRDAs. This shows the DomSum framework can leverage context to compensate the summaries.
%Even without context, both of the topic models performs competively to the supervised methods in terms of F1 score.

\begin{spacing}{0.9}
\begin{table}
%    \hspace{0.2cm}
    {\scriptsize
    \setlength{\baselineskip}{0pt}
    \begin{tabular}{|p{0.7in}|p{0.32in}|p{0.32in}|p{0.32in}|p{0.32in}|p{0.32in}|}
        \hline
\textbf{ }&\multicolumn{5}{|c|}{\textbf{System Clusterings}}\\
\hline
\textbf{ }&\multicolumn{3}{|c|}{\textbf{R-1}}&\textbf{R-2}&\textbf{R-SU4}\\
%\cline{3-7}
\hline
 &PREC&REC&F1&F1&F1\\
    {\bf Baselines}& & & & &\\
  Longest DA&17.06  &11.64  &13.84& 2.76&3.34\\
  Prototype DA&18.14  &10.11  &12.98& 2.84&3.09\\
  \hline\hline
    {\bf Supervised Methods} & & &  & &\\
    CRF&46.97  &15.25  &23.02 & 6.09&9.11\\
    SVM&39.05  & 18.45  & 25.06&  6.11& 9.82\\
  \hline\hline
     {\bf Our Approach} & & &  & &\\
     {\bf 5 topics} & & &  & &\\
    LocalLDA &25.57  &16.57  &20.11 &4.03 &5.87\\
	\ \ \ \ + {\it context} &\ \ \ 20.68  &\ \ \ 25.96  &\ \ \ 23.02 &\ \ \ 3.09 &\ \ \ 4.48\\
    STM & 24.15  &17.82  &20.51 &4.03 &5.69\\
    \ \ \ \ + {\it context} &\ \ \ 20.64  &\ \ \ \textit{\textbf{30.03}}  &\ \ \ {\bf 24.47} &\ \ \ 3.59 &\ \ \ 4.76\\
    \hline
    {\bf 10 topics} & & &  & &\\
    LocalLDA & 25.98  &15.94  &19.76 &3.59 &4.41\\
    \ \ \ \ +  {\it context} &\ \ \ 23.98  &\ \ \ 21.92  &\ \ \ 22.90 &\ \ \ 3.45 &\ \ \ 4.10\\
    STM & 26.32  &19.14  & 22.16 &{\bf 4.07} &{\bf 5.88}\\
    \ \ \ \ +  {\it context} &\ \ \ 22.50  &\ \ \ 28.40  &\ \ \ 25.11 &\ \ \ 3.43 &\ \ \ 4.15\\

\hline
    \end{tabular}
    }
    \vspace{-0.2cm}
    \caption{\footnotesize  ROUGE-1 (R-1), ROUGE-2 (R-2) and ROUGE-SU4 (R-SU4) scores for our proposed token-level summarization approaches, compared with two baselines, supervised methods. --- all use System Clusterings}
    \label{tab:rouge comparison results_system}
\end{table}
\end{spacing}

\vspace{-0.2cm}
\subsection{System Clusterings}
\vspace{-0.2cm}
Results using the {\bf System Clusterings} (Table~\ref{tab:rouge comparison results_system}) present similar findings, though all
of the system and baseline scores are lower. By adding context information, the token-level summarization approaches based on fine-grained topic models compare favorably to the supervised methods in F scores, and also get the best ROUGE-1 recalls.

\vspace{-0.2cm}
\subsection{Sample System Summaries}
\vspace{-0.2cm}
To better exemplify the summaries generated by different systems,
sample output for each method is shown in Table~\ref{tab:sample output}.  We see from the table that
utterance-level extractive summaries (Longest DA, Prototype DA,
TMM) make more coherent but still far from concise and compact
abstracts. On the other hand, the supervised methods (SVM, CRF) that
produce token-level extracts better identify the overall content of
the decision abstract.  Unfortunately, they require
human annotation in the training phase. In comparison, the output of fine-grained topic models can cover the most useful information.

\begin{table}
    \hspace{-0.1cm}
    {\scriptsize
    \setlength{\baselineskip}{0pt}
    \begin{tabular}{|l|}
        \hline
    %{\bf Four DAs in the cluster}  \\

  {\bf DRDA (1)}: I think if we can if we can include them at not too\\ much extra cost, then I'd put them in,\\
  {\bf DRDA (2)}: Uh um we we're definitely going in for voice\\ recognition as well as LCDs, mm.\\
  {\bf DRDA (3)}: So we've basically worked out that we're going\\ with a simple battery,\\
  
  %{\bf context DA (1)}:using the simple battery will be a safer option\\ as compared to the kinetic energy one, I mean,\\
  {\bf context DA (1)}:So it's advanced integrated circuits?\\
  {\bf context DA (2)}:the advanced chip\\
  {\bf context DA (3)}: and a curved on one side case which is folded\\ in on itself , um made out of rubber\\
  \hline
  \hline

  {\bf Decision Abstract}: It will have voice recognition, use a simple\\ battery, and contain an advanced chip.\\
  \hline
  \hline

  {\bf Longest DA \& Prototype DA:}
  Uh um we we're definitely going\\ in for voice recognition as well as LCDs, mm.\\
%  {\bf Prototype DA:}\\
%  Uh um we we're definitely going in for voice recognition as well\\ as LCDs, mm.\\
  {\bf TMM:}
  I think if we can if we can include them at not too much\\ extra cost, then I'd put them in,\\
  
  {\bf SVM:} cost voice recognition simple battery\\
  {\bf CRF:} voice recognition battery\\
  
%  {\bf localLDA:}\\ 
%  extra cost, definitely going voice recognition, simple battery\\
  {\bf STM:}
  extra cost, definitely going voice recognition LCDs, \\simple battery\\
%  {\bf localLDA+context:}\\
%  cost, company, advanced, going voice recognition, simple battery,\\ chip, curved case rubber\\
  {\bf STM + context:}
  cost, company, advanced integrated circuits, going \\voice recognition, simple battery, advanced chip, curved case rubber\\

%  {\bf OneTopic:}\\
%  So we've basically worked out that we're going\\ with a simple battery,\\  
%  {\bf MultiTopic:}\\
%  So we've basically worked out that we're going\\ with a simple battery,\\
  
  \hline
    \end{tabular}
    }
    \vspace{-0.2cm}
    \caption{\footnotesize Sample system outputs by different methods are in the third cell (methods' names are in bold). First cell contains three DRDAs supporting the decision in the second cell and three adjacent DAs of them.}
    \label{tab:sample output}
\end{table}

\section{Conclusion}
\vspace{-0.2cm}
We propose a token-level summarization framework based on topic models and show that modeling topic structure at the utterance-level is better at identifying relevant words and phrases than document-level models. The role of context is also studied and shown to be able to identify additional summary-worthy words. %Our immediate future work involves a continuing investigation into modeling the discourse structure within the dialogue acts which may lead to structured summarization.
%For further work, we will investigate other speech features, such as acoustic or prosodic features. 
%We also will use the output of this approach for further generating abstractive summaries.

\vspace*{0.5mm}
\begin{small}
\noindent
{\bf Acknowledgments}
This work was supported in part by National Science Foundation Grants
IIS-0968450 and IIS-1111176, and by a gift from Google.
\end{small}

{\scriptsize
\bibliographystyle{acl2012}

}
\end{document}